\pdfoutput=1

\documentclass[11pt]{article}

\usepackage{coling}

\usepackage{subcaption}
\usepackage{tikz}
\usepackage{pgfplots}
\usepackage{times}
\usepackage{latexsym}

\usepackage[T1]{fontenc}

\usepackage[utf8]{inputenc}

\usepackage{microtype}

\usepackage{inconsolata}

\usepackage{graphicx}

\usepackage{latexsym}
\usepackage{array}
\usepackage{booktabs}
\usepackage{siunitx}
\usepackage{amsmath,amssymb,amsthm}
\usepackage{multirow}

\usepackage{caption}
\usepackage{subcaption}

\title{BabyLMs for isiXhosa:\\Data-Efficient Language Modelling in a Low-Resource Context}

\author{Alexis Matzopoulos \,\,\, Charl Hendriks \,\,\, Hishaam Mahomed \,\,\, Francois Meyer\\
  Department of Computer Science \\
  University of Cape Town \\
  \texttt{\{mtzale001, hmhis005, hndcha033\}@myuct.ac.za, francois.meyer@uct.ac.za}}

\begin{document}
\maketitle
\begin{abstract}
The BabyLM challenge called on participants to develop sample-efficient language models. Submissions were pretrained on a fixed English corpus, limited to the amount of words children are exposed to in development (<100m). The challenge produced new architectures for data-efficient language modelling, which outperformed models trained on trillions of words. This is promising for low-resource languages, where available corpora are limited to much less than 100m words. In this paper, we explore the potential of BabyLMs for low-resource languages, using the isiXhosa language as a case study. We pretrain two BabyLM architectures, ELC-BERT and MLSM, on an isi\-Xhosa corpus. They outperform a vanilla pretrained model on POS tagging and NER, achieving notable gains (+3.2 F1) for the latter. In some instances, the BabyLMs even outperform XLM-R. Our findings show that data-efficient models are viable for low-resource languages, but highlight the continued importance, and lack of, high-quality pretraining data. Finally, we visually analyse how BabyLM architectures encode isi\-Xhosa. 


\end{abstract}

\section{Introduction}

Large language models (LLMs) are trained on trillions of words \citep{touvron2023llama2openfoundation}. Humans are much more efficient language learners -- children are exposed to less than 100 million words of speech/text by age 13 \citep{gilkerson-etal-2017}. This mismatch motivated the establishment of the BabyLM challenge \citep{warstadt-etal-2023-findings}, a shared task in which participants were invited to propose data-efficient language modelling techniques. Submissions were pretrained on a fixed corpus of developmentally plausible English (e.g. child-directed speech, educational content) and ranked according to performance on natural language understanding (NLU) benchmarks. 

The top submissions comfortably outperformed standard Transformer-based \citep{vaswani2023attentionneed} models 
pretrained on the same fixed corpus, even surpassing state-of-the-art pretrained language models (PLMs) trained on orders of magnitude more data. The main aims of the BabyLM challenge was to build cognitively plausible models of language acquisition and enable compute-limited language modelling research \citep{warstadt-etal-2023-findings}. In this paper, we investigate an additional opportunity arising from the shared task: its potential to improve LMs for low-resource languages.

\begin{table}[t]
\centering
\begin{tabular}{lccc} 
\toprule
 \textbf{Model} & POS & NER & NTC \\ 
  \midrule
 \multicolumn{4}{l}{\textbf{Pretrained on 13m isiXhosa words}} \\
 \midrule
 RoBERTa & 87.0$_{\pm0.1}$ & 85.4$_{\pm0.4}$ & \textbf{97.6}$_{\pm0.5}$ \\
 MLSM & 87.4$_{\pm0.1}$ & 87.0$_{\pm0.4}$ & 95.4$_{\pm0.2}$ \\
 ELC-BERT& \textbf{87.7}$_{\pm0.5}$  & \textbf{88.6}$_{\pm0.6}$ & 95.0$_{\pm0.3}$ \\ 
 \midrule
 \multicolumn{4}{l}{\textbf{Massively multilingual pretraining}} \\
  \midrule
 XLM-R & 88.1  & 88.1 & 89.2\\
 Afro-XLMR & \underline{\textbf{88.7}} & 89.9 & 97.2 \\
 Nguni-XLMR & 88.3 & \underline{\textbf{90.4}} & \underline{\textbf{98.2}}\\
\bottomrule
\end{tabular}
\caption{BabyLM performance on isiXhosa tasks, compared to a RoBERTa baseline trained from scratch and three large-scale multilingual PLMs. We \textbf{boldface} best per-category performance and \underline{underline} best overall.}
\label{Table2}
\end{table}

BabyLMs aim to optimise performance on a limited training budget. For the BabyLM challenge, this was simulated by creating a constrained English corpus. For low-resource languages, such constraints represent the reality of their NLP resources. Most languages do not have publicly available corpora consisting of trillions of words, so out of necessity they operate on a limited training budget. The data-efficiency of BabyLMs therefore presents a promising opportunity to achieve real-world performance gains for certain languages.

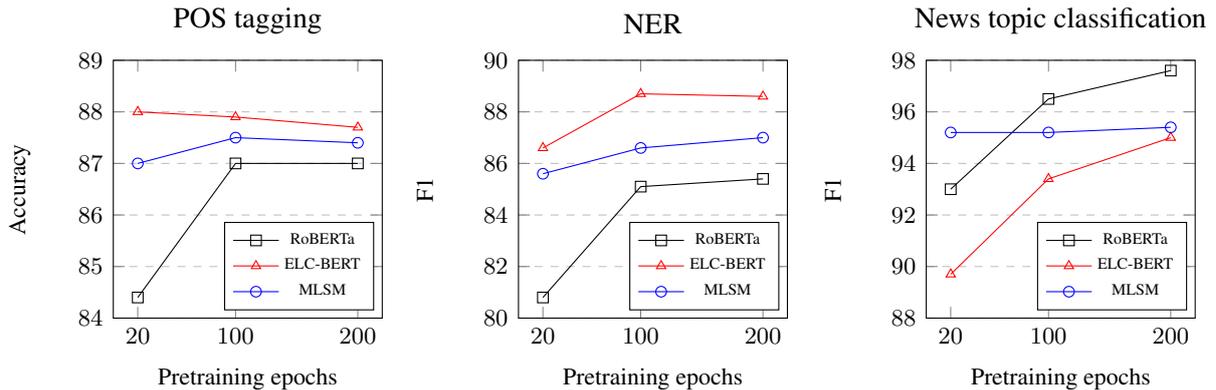
\begin{figure*}
\centering
\hfill
\begin{subfigure}[b]{0.32\textwidth}
\begin{tikzpicture}
\begin{axis}[
    width=\textwidth,
    height=5cm,
    xlabel={Pretraining epochs},
    ylabel={Accuracy},
    xmin=0, xmax=220,
    ymin=84, ymax=89,
    xtick={20,100,200},
    ytick={84,85,86,87,88,89},
    legend pos=south east,
    ymajorgrids=true,
    grid style=dashed,
    title={POS tagging},
    legend style={font=\tiny},
    label style={font=\footnotesize},
    tick label style={font=\footnotesize}
]

\addplot[
    color=black,
    mark=square,
    ]
    coordinates {
    (20,84.4)(100,87.0)(200,87.0)
    };
    
\addplot[
    color=red,
    mark=triangle,
    ]
    coordinates {
    (20,88.0)(100,87.9)(200,87.7)
    };
    
\addplot[
    color=blue,
    mark=o,
    ]
    coordinates {
    (20,87.0)(100,87.5)(200,87.4)
    };
    
\legend{RoBERTa,ELC-BERT,MLSM}
\end{axis}
\end{tikzpicture}
\end{subfigure}
\hfill
\begin{subfigure}[b]{0.32\textwidth}
\begin{tikzpicture}
\begin{axis}[
    width=\textwidth,
    height=5cm,
    xlabel={Pretraining epochs},
    ylabel={F1},
    xmin=0, xmax=220,
    ymin=80, ymax=90,
    xtick={20,100,200},
    ytick={80,82,84,86,88,90},
    legend pos=south east,
    ymajorgrids=true,
    grid style=dashed,
    title={NER},
    legend style={font=\tiny},
    label style={font=\footnotesize},
    tick label style={font=\footnotesize}
]

\addplot[
    color=black,
    mark=square,
    ]
    coordinates {
    (20,80.8)(100,85.1)(200,85.4)
    };
    
\addplot[
    color=red,
    mark=triangle,
    ]
    coordinates {
    (20,86.6)(100,88.7)(200,88.6)
    };
    
\addplot[
    color=blue,
    mark=o,
    ]
    coordinates {
    (20,85.6)(100,86.6)(200,87.0)
    };
    
\legend{RoBERTa,ELC-BERT,MLSM}
\end{axis}
\end{tikzpicture}
\end{subfigure}
\hfill
\begin{subfigure}[b]{0.32\textwidth}
\begin{tikzpicture}
\begin{axis}[
    width=\textwidth,
    height=5cm,
    xlabel={Pretraining epochs},
    ylabel={F1},
    xmin=0, xmax=220,
    ymin=88, ymax=98,
    xtick={20,100,200},
    ytick={88,90,92,94,96,98},
    legend pos=south east,
    ymajorgrids=true,
    grid style=dashed,
    title={News topic classification},
    legend style={font=\tiny},
    label style={font=\footnotesize},
    tick label style={font=\footnotesize}
]

\addplot[
    color=black,
    mark=square,
    ]
    coordinates {
    (20,93.0)(100,96.5)(200,97.6)
    };
    
\addplot[
    color=red,
    mark=triangle,
    ]
    coordinates {
    (20,89.7)(100,93.4)(200,95.0)
    };
    
\addplot[
    color=blue,
    mark=o,
    ]
    coordinates {
    (20,95.2)(100,95.2)(200,95.4)
    };
    
\legend{RoBERTa,ELC-BERT,MLSM}
\end{axis}
\end{tikzpicture}
\end{subfigure}
\caption{Downstream task performance for model checkpoints at different stages of pretraining.}
\label{fig:nlu-results}
\end{figure*}

To investigate BabyLMs in a low-resource context we turn to isiXhosa, a South African language with over 22 million speakers \citep{eberhard-etal-2019-ethnologue}. We pretrain two of the top BabyLM submissions, Every Layer Counts BERT (ELC-BERT) \citep{georges-gabriel-charpentier-samuel-2023-layers} and Masked Latent Semantic Modeling (MLSM) \citep{berend-2023-masked}, for isiXhosa. We evaluate on isiXhosa NLU tasks and compare performance to a baseline RoBERTa architecture \citep{liu2019robertarobustlyoptimizedbert} pretrained on the same isiXhosa corpus. 

Our results confirm the potential of data-efficient architectures for low-resource languages, with both BabyLMs obtaining performance gains over the RoBERTa baseline on POS tagging and NER. ELC-BERT proves especially effective, even rivalling one of our \emph{skylines} (large-scale existing PLMs for isiXhosa). Unlike in the BabyLM challenge, our models do not outperform the best skylines, which we attribute to a lack of developmentally plausible data for isiXhosa. In summary, while our results indicate that low-resource gains are available from architectural innovations, they also highlight the continued need to develop higher-quality datasets for low-resource languages.

\section{Background}

\subsection{PLMs for isiXhosa}

Pretraining corpora for isiXhosa are limited to 20m words \citep{xue-etal-2021-mt5}. This is greater availability than most languages, but still two orders of magnitude less than even early PLMs \citep{devlin-etal-2019-bert}. As for other low-resource languages, multilingual modelling has improved performance for isiXhosa NLU. 
IsiXhosa is included in \textbf{XLM-R} \cite{conneau-etal-2020-unsupervised}, a masked language model (MLM) pretrained on 100 languages.
Two previous works improved performance for isi\-Xhosa by adapting XLM-R through continued pretraining. \textbf{Afro-XLMR} \citep{alabi-etal-2022-adapting} adapts XLM-R for 23 African languages, including isiXhosa. \textbf{Nguni-XLMR} \cite{meyer-etal-2024-nglueni} narrows the linguistic scope by adapting XLM-R for the four Nguni languages (isiXhosa, isiZulu, isiNdebele, Siswati), the closest linguistic relatives of isiXhosa. 

\begin{figure*}[t] 
  \centering
  \begin{subfigure}[b]{0.48\textwidth}
    \centering
    \includegraphics[trim={0 0 0 1cm},clip,width=\textwidth,keepaspectratio]{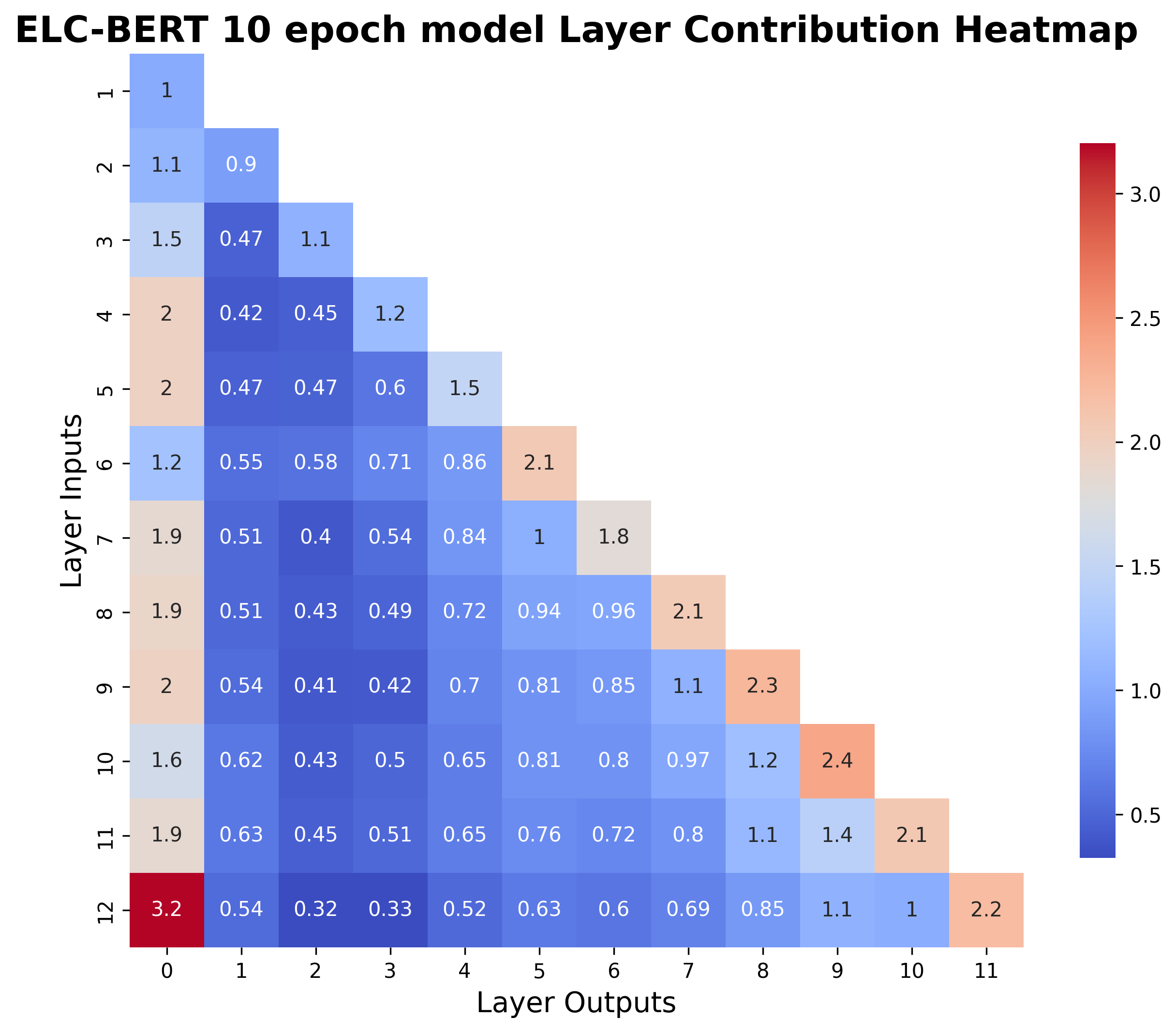}
    \caption{After 10 epochs of training.}
    \label{fig:10_epochs}
  \end{subfigure}
  \hfill
  \begin{subfigure}[b]{0.48\textwidth}
    \centering
    \includegraphics[trim={0 0 0 1cm},clip,width=\textwidth,keepaspectratio]{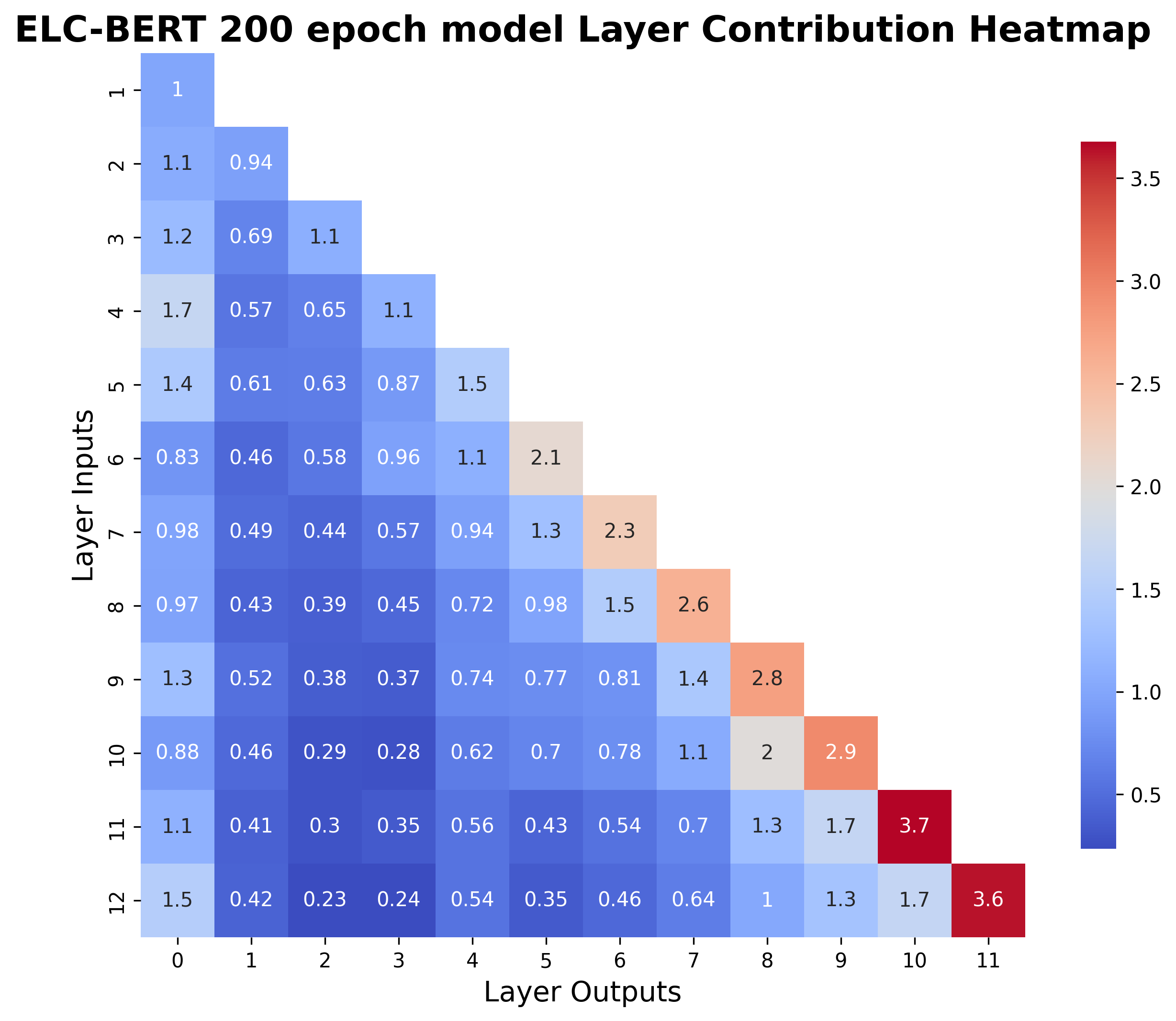}
    \caption{After 200 epochs of training.}
    \label{fig:200_epochs}
  \end{subfigure}
  \caption{Layer contribution heatmaps of isiXhosa ELC-BERT at different stages of pretraining.}
  \label{fig:heatmaps_side_by_side}
\end{figure*}

\subsection{BabyLM Architectures}

The BabyLM challenge hosted three competition tracks, corresponding to different data restrictions.
The \emph{Small} and \emph{Strict-Small} tracks were respectively limited to 100m and 10m words for pretraining, while the \emph{Loose} track allowed non-linguistic data. 
ELC-BERT \citep{georges-gabriel-charpentier-samuel-2023-layers} won both the \emph{Small} and \emph{Strict-Small} tracks, outperforming skyline models Llama2 \citep{touvron2023llama2openfoundation} and RoBERTa-base \citep{liu2019robertarobustlyoptimizedbert}. MLSM \citep{berend-2023-better} was runner-up in the \emph{Strict-Small} track. The \emph{Strict-Small} data restriction (10m words) most closely aligns with the size of publicly available corpora for isiXhosa, which is why we chose the top models from this category.

\subsubsection{Every Layer Counts BERT (ELC-BERT)}
ELC-BERT \cite{georges-gabriel-charpentier-samuel-2023-layers} adapts LTG-BERT \cite{samuel-etal-2023-trained}, an architecture designed to optimise pretraining on small corpora. ELC-BERT modifies residual connections to selectively weigh outputs from previous layers. Each layer's input is a combination of outputs from previous layers, weighted by learnable layer-specific weights. This is in contrast to standard residual connections, where the input is an equally weighted sum of all preceding outputs. The added expressivity of ELC-BERT, which allows the model to dynamically weigh how preceding layers are incorporated into computations, enables more sample-efficient learning.

\subsubsection{Masked Latent Semantic Modeling (MLSM)}
MLSM \cite{berend-2023-masked, berend-2023-better} is an alternative to standard masked language modeling. Instead of tasking the model with predicting specific tokens, which can be challenging given limited training data, the model is trained to predict broader semantic categories. For example, if the model is tasked to predict the masked word ``barbecue'', it would generate predictions towards the semantic attributes associated with the word (e.g. ``food'', ``outdoors'', ``fire'').
MLSM uses a teacher model to determine latent semantic distributions for masked tokens, via sparse coding of their hidden representations. The final model is then a student model, trained to predict these latent semantic distributions rather than the exact identities of masked tokens.

\section{Experimental Setup}

\paragraph{Pretraining}
Our BabyLMs and baseline are pretrained on the WURA isi\-Xhosa corpus \cite{oladipo-etal-2023-better}, which is a compiled by filtering mC4 \cite{xue-etal-2021-mt5} to remove noise. 
The isi\-Xhosa dataset contains 13m words, similar in size to BabyLM \emph{Strict-Small}. 
Our models are trained for 200 epochs on a Tesla V100 GPU. We detail our training process in \autoref{sec:appendixA}.

\begin{figure*}[t]
  \centering
  \includegraphics[width=\textwidth,keepaspectratio]{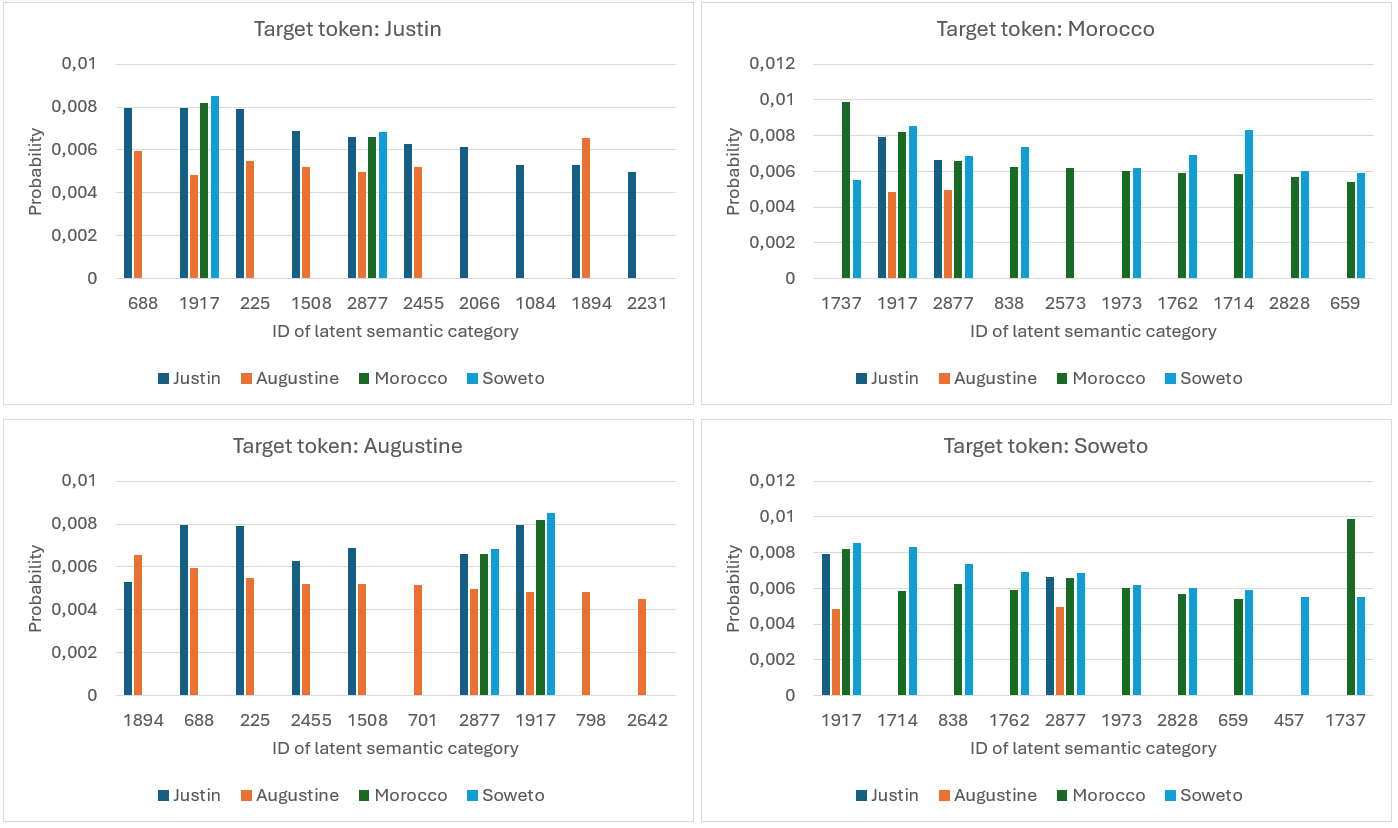}
  \caption{Top 10 semantic categories predicted by isiXhosa MLSM for named entities (sampled from MasakhaNER).}
  \label{NER}
\end{figure*}

\paragraph{Evaluation}
We evaluate on three isiXhosa datasets --  MasakhaPOS \cite{dione-etal-2023-masakhapos} for POS tagging, MasakhaNER \cite{adelani-etal-2022-masakhaner} for NER, and MasakhaNEWS \cite{adelani-etal-2023-masakhanews} for news topic classification (NTC). 
Test set results are averaged across 5 finetuning runs. 

\section{Results}

\autoref{Table2} presents our results. Both BabyLMs outperform the baseline on POS and NER, achieving large gains for NER (+3.2 F1 for ELC-BERT and +1.6 F1 for MLSM). As in the original shared task, ELC-BERT is the top-performing BabyLM. ELC-BERT demonstrates superior efficiency in both data utilisation (shown in \autoref{fig:nlu-results}) and computate requirements (its pretraining time is 70\% faster than MLSM).
The BabyLMs fail to outperform RoBERTa on NTC. We attribute this to topic classification being an easier task than POS and NER, so data-efficiency is less critical. In fact, our RoBERTa baseline even outperforms two skylines on NTC, reaffirming previous findings that pretraining from scratch is sufficient for the simpler task of NTC \citep{ogueji-etal-2021-small, dossou-etal-2022-afrolm}. 
We also posit that the architectures of ELC-BERT and MLSM are more suitable for word-level tasks than sequence-level tasks (discussed in \autoref{sec5}).

ELC-BERT outperforms one skyline, XLM-R, on two tasks. Unlike in the shared task, our models do not outperform the top skylines. We attribute this to an important difference between our setup and the shared task -- the quality of pretraining data. The WURA corpus does not match the quality of the BabyLM data, which was curated to include developmentally plausible text (e.g. child-directed speech, educational content). The previous success of these models in English is due to a combination of modelling innovations and extremely high-quality pretraining data, which is lacking for low-resource languages like isiXhosa.

\section{Analysis} \label{sec5}

The BabyLMs studied in this paper achieve data-efficiency by augmenting the standard MLM architecture. We now analyse how their unique architectural innovations encode the isiXhosa language.

\paragraph{ELC-BERT} The residual connections of ELC-BERT learn to selectively weigh the output of previous layers. We visualise learned weights in \autoref{fig:heatmaps_side_by_side}, comparing early pretraining to complete pretraining (intermediary stages are visualised in \autoref{fig:heatmaps_side_by_side2} in the appendix). The weighting exhibits significant deviations from a standard Transformer layer (which assigns equal weight to all preceding outputs). In early stages of pretraining, the model is biased to the embedding layer and immediately preceding layers.
As pretraining progresses, the model reduces its reliance on embeddings in favour of immediately preceding layers, but still assigns more weight to the embedding layer than the BabyLM ELC-BERT submission.
We posit that this emphasis on the embedding layer underlies ELC-BERT's performance gains on POS tagging and NER, since embeddings encode information about word-level syntactic roles \citep{tenney-etal-2019-bert}.

\paragraph{MLSM}

During pretraining, MLSM predicts the latent semantic categories of masked tokens. 
To inspect the semantic distribution learned by the model, we extract the predictions for masked named entities in sentences sampled from MasakhaNER. 
\autoref{NER} shows the top 10 semantic categories (each corresponding to an index) 
assigned to four named entities. For each target word, we also plot the probabilities produced for the other words to compare distributions.  
In our sampled sentences, two of the words (\emph{Justin} and \emph{Augustine}) are names of persons, while the other two (\emph{Morocco} and \emph{Soweto}) are names of locations. The plots demonstrate more semantic overlap between the same types of named entities. The names of persons have seven overlapping semantic categories, while the names of locations have nine overlapping categories. Between the two named entity types, only two semantic categories overlap. This pattern indicates that MLSM effectively encodes the semantic properties of these named entities, to which we attribute its NER performance gains. We present a similar analysis for target words across POS tags in \autoref{sec:appendixB}.

\section{Conclusion}



This study explored the potential of two architectures from the BabyLM challenge, ELC-BERT and MLSM, to benefit low-resource languages. Comparing our findings to those of the BabyLM challenge, we draw three main conclusions. 
Firstly, the gains obtained by isi\-Xhosa BabyLMs show that the sample-efficiency sought by the BabyLM challenge can prove effective in real low-resource settings. 
Secondly, ELC-BERT once again emerges as the most data-efficient solution, even outperforming massively multilingual PLMs. 
Lastly, the fact that our BabyLMs do not outperform all skylines shows that the absence of high-quality corpora for isi\-Xhosa poses a barrier to further gains. The findings of the BabyLM challenge can be attributed to both architectural innovations and  specifically curated pretraining data. The BabyLM pretraining corpus includes child-directed speech, educational video subtitles, and articles from Simple Wikipedia (an edition of Wikipedia written in simplified English, using shorter sentences and common words). Such high-quality, developmentally plausible data is not publicly available for isi\-Xhosa. Our results show that this limits the potential of BabyLMs for low-resource languages.

More generally, this work unites two directions of research -- cognitively plausible modelling and NLP for low-resource languages. We hope more researchers pursue work at the intersection of these two subfields, since they share the goal of improving data-efficiency in the era of scaling. 

\section{Limitations}

Our study focussed on a single language, isiXhosa, so our findings might not generalise to other low-resource languages. We chose isiXhosa because its data availability was well suited to our study. Publicly available pretraining corpora for isiXhosa are similar in size to the BabyLM \emph{Strict-Small} corpus. In terms of downstream evaluation data, isiXhosa also has sufficient NLU datasets available to allow evaluation across sequence labelling and sequence classification tasks. The BabyLM challenge evaluated submissions across many more tasks than we did, some of which are much more challenging than our isiXhosa evaluation tasks. Ideally, one would evaluate our isiXhosa BabyLMs on datasets that test more aspects of language competence. This would reveal further insights into the value of BabyLM architectures compared to standard baselines and/or skylines, which might not align with our current findings. We hypothesise that more complex evaluation tasks would further highlight the value of BabyLMs over standard Transformer baselines, but due to the lack of additional isiXhosa evaluation datasets we are unable to test this.

\bibliography{custom, anthology_0, anthology_1, new}

\appendix

\section{Training Details} \label{sec:appendixA}

For pretraining, we use the training scripts accompanying the BabyLM submissions, and use their hyperparameter settings for the \emph{Strict-Small} track as a starting point. We pretrain our BabyLMs and RoBERTa baseline for 200 epochs of the isiXhosa WURA corpus. Our hyperparameter settings are listed in  \autoref{Table1}.

\begin{table}[h!]
\centering
\begin{tabular}{c c c c c} 
 \toprule
 \textbf{Model} & \textbf{LR} & \textbf{SL} & \textbf{H} & \textbf{BS}\\
 \midrule
 RoBERTa & $5e^{-5}$ & 512 & 12 & 8 \\
 ELC-BERT & $5e^{-4}$ & 128 & 12 & 128 \\
 MLSM (teacher) & $1e^{-4}$ & 128 & 12 & 64 \\
 MLSM (student) & $1e^{-4}$ & 128 & 12 & 64 \\ 
 \bottomrule
\end{tabular}
\caption{Pretraining hyperparameters (Learning Rate, Sequence Length, Hidden layers, Batch Size)}
\label{Table1}
\end{table}

\paragraph{ELC-BERT pretraining}
Due to computational constraints, we trained our ELC-BERT model for 200 epochs, instead of the 2000 epochs of the BabyLM submission. Regardless, downstream performance for POS tagging and NER does plateau by 200 epochs (\autoref{fig:nlu-results}). Besides the number of epochs, we made two changes to the hyperparameter settings of the ELC-BERT submission \citep{georges-gabriel-charpentier-samuel-2023-layers}. Firstly, we used a batch size of 128 (instead of 256) due to computational constraints. Secondly, the original learning rate ($1e^{-2}$) produced an unstable training loss, so after some experimentation we settled on a learning rate of $5e^{-4}$. 

\paragraph{MLSM  pretraining} We trained the teacher and student model from scratch on the WURA dataset, keeping the same hyperparameters as the MLSM submission \citep{berend-2023-better}. Our teacher model is based on the BERT-base-cased architecture\footnote{\url{https://huggingface.co/google-bert/bert-base-cased}} and is trained using a standard masked language modelling objective. We used the teacher model hidden layers to create a semantic dictionary for the student model. The student model is also based on the BERT-base-cased architecture, but is trained to predict semantic categories instead of masked tokens.

\paragraph{Finetuning}
We use the finetuning scripts provided by the MasakhaPOS \cite{dione-etal-2023-masakhapos}, MasakhaNER \cite{adelani-etal-2022-masakhaner}, and MasakhaNEWS \cite{adelani-etal-2023-masakhanews} datasets where possible, and adapt them for ELC-BERT. Each model is fine-tuned for 20 epochs per task, using the default hyperparameters provided in the respective dataset fine-tuning scripts. For each task, we perform 5 finetuning runs using different random seeds. We report the averages and standard deviations over these runs in \autoref{Table2}.

\begin{figure*}[t!] 
  \centering
  \begin{subfigure}[b]{0.48\textwidth}
    \centering
    \includegraphics[trim={0 0 0 1cm},clip,width=\textwidth,keepaspectratio]{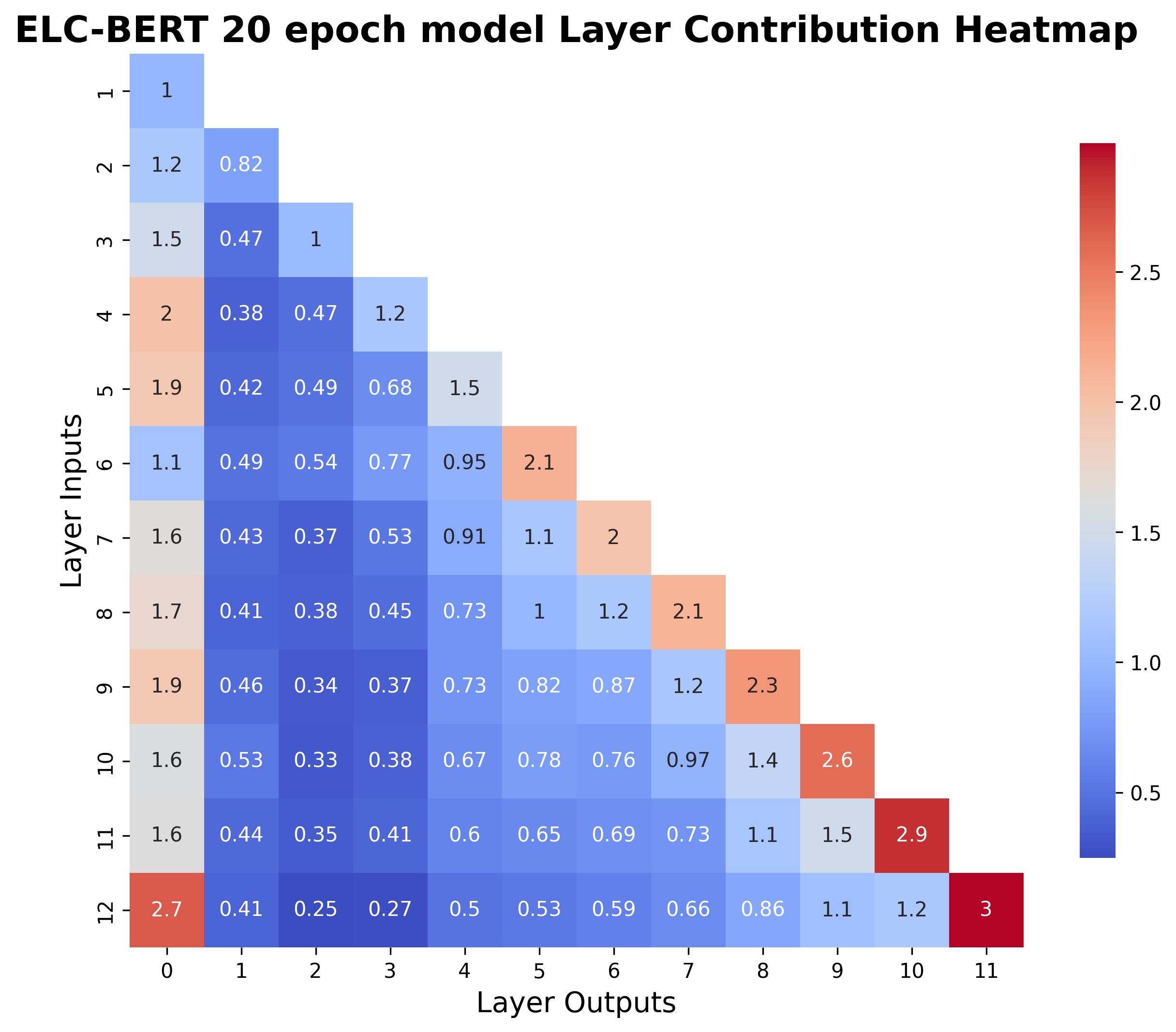}
    \caption{After 20 epochs of training.}
    \label{fig:20_epochs}
  \end{subfigure}
  \hfill
  \begin{subfigure}[b]{0.48\textwidth}
    \centering
    \includegraphics[trim={0 0 0 1cm},clip,width=\textwidth,keepaspectratio]{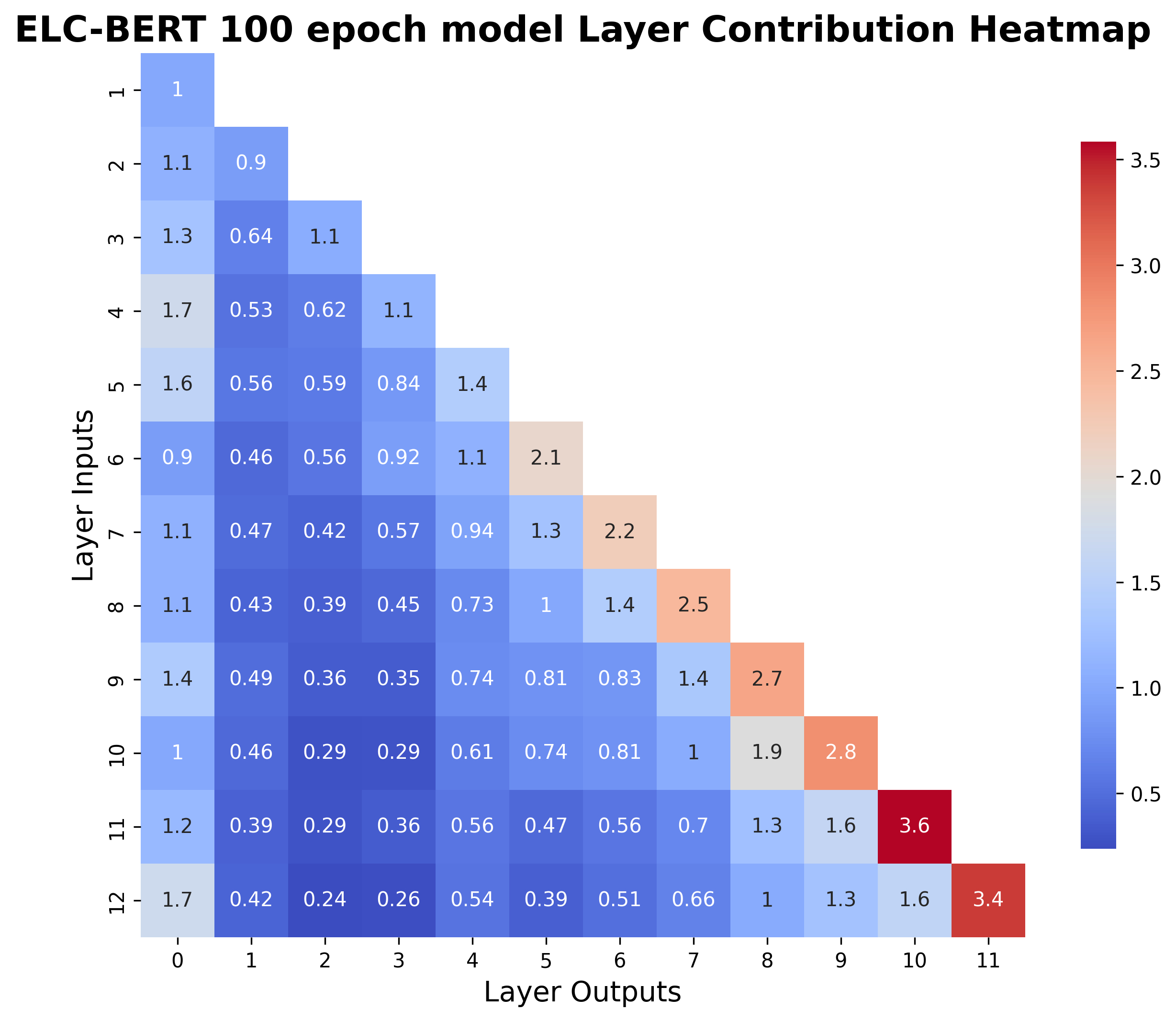}
    \caption{After 100 epochs of training.}
    \label{fig:100_epochs}
  \end{subfigure}
  \caption{Layer contribution heatmaps of isiXhosa ELC-BERT at different stages of pretraining.}
  \label{fig:heatmaps_side_by_side2}
\end{figure*}
\begin{figure*}[t!]
  \centering
  \includegraphics[width=\textwidth,keepaspectratio]{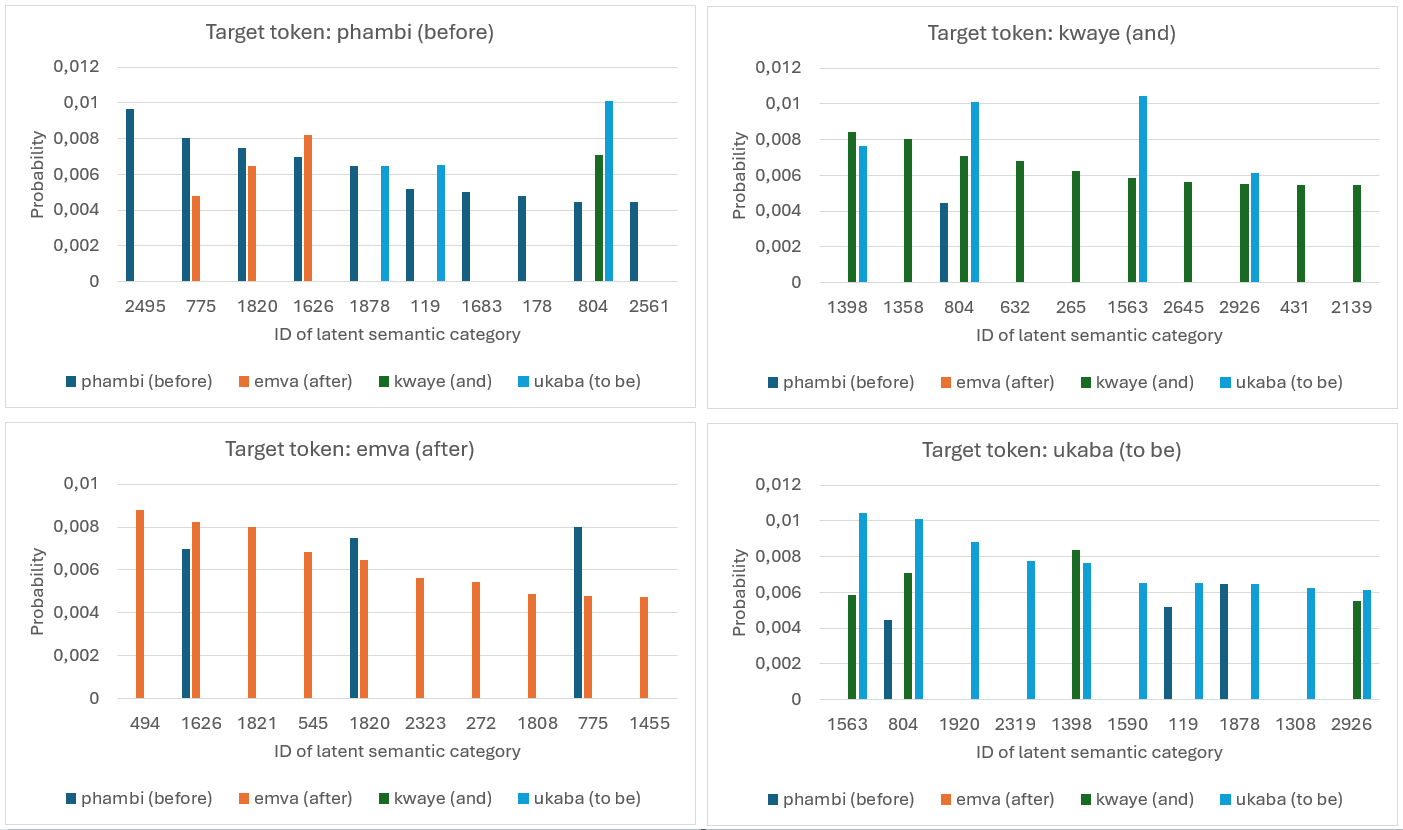}
  \caption{Top 10 semantic categories predicted by isiXhosa MLSM for target words (sampled from MasakhaPOS).}
  \label{POS}
\end{figure*}

\section{MLSM Analysis}\label{sec:appendixB}

The predictions shown in \autoref{NER} are obtained by masking the target words in sentences sampled from MasakhaNER. We conduct a similar analysis for target words with different POS tags, sampling sentences from MasakhaPOS. \autoref{POS} shows the top 10 semantic categories assigned to four words with different parts of speech. Two of the words (\emph{phambi} and \emph{emva}) are adpositions, while the other two (\emph{kwaye} and \emph{ukaba}) are conjunctions. The plots demonstrate less semantic overlap between same POS tags than named entity types. The adpositions have three overlapping semantic categories, while the conjunctions share four overlapping categories. Between the different POS tags, there is still minimal overlap: \emph{phambi} shares one category with \emph{kwaye} and three categories with \emph{ukaba}, while \emph{emva} shows no overlap with either conjunction. We attribute this pattern to the broader and less interchangeable nature of POS tags compared to named entities, making them less suited to MLSM's strengths. The reduced semantic overlap, compared to named entities, might be why MLSM's effectiveness varies across linguistic tasks. This aligns with the results shown in \autoref{Table2}, where MLSM's performance gains for POS tagging show a narrower margin over the baseline compared to the improvements in NER.

\end{document}